\documentclass{article}

\usepackage{arxiv}
\usepackage[toc,page]{appendix}
\usepackage{caption}
\usepackage{subcaption}
\usepackage[utf8]{inputenc} 
\usepackage[T1]{fontenc}  
\usepackage{hyperref}    
\usepackage{url}      
\usepackage{booktabs}    
\usepackage{amsfonts}    
\usepackage{nicefrac}    
\usepackage{microtype}   
\usepackage{lipsum}		
\usepackage{graphicx}
\usepackage{natbib}
\usepackage{doi}
\graphicspath{{figures/}} 

\title{Sparsely  ensembled  convolutional  neural  network  classifiers  via  reinforcement learning}

\date{} 					

\author{ \href{https://orcid.org/0000-0002-2493-839X}{\includegraphics[scale=0.06]{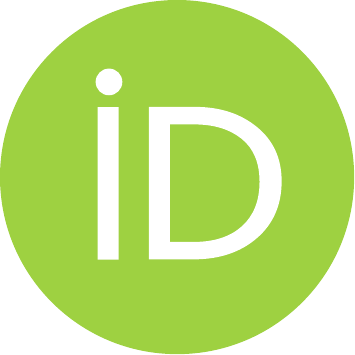}\hspace{1mm}Roman O.~Malashin $ {}^{1,2} $}\thanks{e-mail: malashinroman@mail.ru} \\
	$ {}^{1} $ Group of neural networks and artificial intelligence,\\
	Pavlov institute of Physiology RAS, \\
	$ {}^{2} $ State University of Aerospace Instrumentation, \\
	Saint-Petersburg, Russia
}

\renewcommand{\shorttitle}{\textit{arXiv} Sparsely ensembelled CNNs}

\hypersetup{
	pdftitle={A template for the arxiv style},
	pdfsubject={ML, AI, CV},
	pdfauthor={R. Malashin},
	pdfkeywords={First keyword, Second keyword, More},
}

\begin{document}
	\maketitle
	
	\begin{abstract}
		We consider convolutional neural network (CNN) ensemble learning with the objective function inspired by least action principle; it includes resource consumption component. We teach an agent to perceive images through the set of pre-trained classifiers and want the resulting dynamically configured system to unfold the computational graph with the trajectory that refers to the minimal number of operations and maximal expected accuracy. The proposed agent’s architecture implicitly approximates the required classifier selection function with the help of reinforcement learning. Our experimental results prove, that if the agent exploits the dynamic (and context-dependent) structure of computations, it outperforms conventional ensemble learning.
	\end{abstract}


	\section{Introduction}
	Ensemble learning is an approach to machine learning, which refers to acquiring a predictor (strong classifier or committee) that has a form of the weighted combination of base models (weak learners). Bagging, boosting, and stacking are well-known ensemble methods with practical applications.
	
	The ensemble learning assumes all weak learners are used for prediction. This violates the principle of minimal energy consumption. We refer to this fundamental as the least action principle \citep{Malashin2019}.  Shelepin et. al showed that the least action principle can be considered as the cognition principle in vision \citep{Shelepin2003, Shelepin2006}. In physics principle of least action states that objects in space follow trajectories that satisfy the minimum of a two-component functional called Action. We adapt this principle: the computational graph should be traversed with a trajectory that satisfies maximum expected accuracy and minimum computational costs. In terms of ensemble learning, if an example is easy, we prefer to rely on the response of just a few weak learners (use short path in the dynamic computational graph), while computationally heavy analysis is justified for hard cases.
	
	Conventional sparse boosting assumes some features may be absent during prediction but does not address the desirability of such “an absence”. A simple but popular approach incorporating the least action principle is decision lists when most of the easy cases can be rejected by the early tests \citep{ViolaJones2001}. Still, the approach is only applicable for binary classification; it lacks many desirable features of dynamic graph configuration \citep{Malashin2019}.
	
	We state the problem of learning sparse ensemble classifiers taking the least action principle into consideration. The problem can be solved via reinforcement learning by teaching an agent to perceive image through the set of CNN classifiers that are learned externally. The final reward of the agent comprises accuracy minus time consumption. In this work, we concentrate on the image classification task, though the approach can be naturally extended in broader areas of data analysis. 
	
\begin{figure}
		\centering
		\includegraphics[scale=.5]{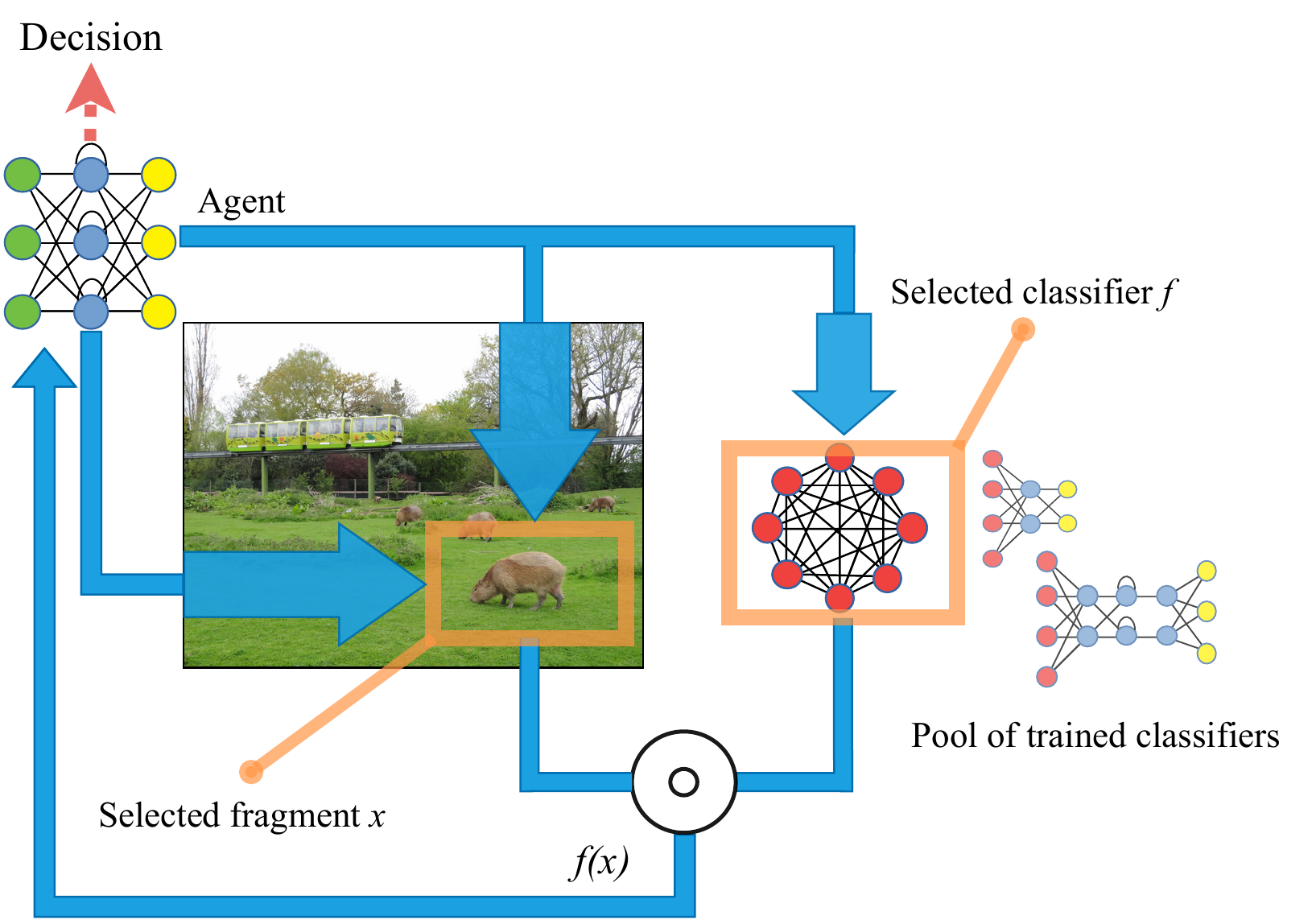}
		\caption{Studied scheme of agent-image interaction through the pool of trained classifiers with visual attention mechanism. Sample photograph by Gareth James [cc-by-sa/2.0] (geograph.org.uk/p/6128774)}
		\label{fig:agentscheme}
	\end{figure}

	The agent’s goal is to learn a policy to optimally select and interpret classifiers on each step with the respect to already revealed particularities of the image. The agent learns a kind of attention mechanism, that can be naturally combined with hard visual attention to select the proper image region for analysis. Figure \ref{fig:agentscheme} depicts the general idea of agent-image interaction through the pool of classifiers with a spatial attention mechanism.
	
	We have found that simultaneous learning of visual attention and classifier selection policies is complicated (due to mutual dependencies of both tasks). In the experimental part, we concentrate on learning classifier selection policy only.
	
	\section{Related works}
	\subsection{Boosting neural networks}
	Classification and regression trees, Haar-wavelets are appropriate weak learners for boosting, but boosting CNN is less studied. One reason is that CNN classifiers provided with enough training data work well without ensemble learning while classification is the main area of boosting. Even more important is that a neural network itself implicitly is an ensemble (where hidden units are weak learners and output unit is an ensemble \citep{Murphy2013}) while being more powerful than the stage-wise additive model (on which conventional boosting relies).
	
	Moghimi and Li \citep{Moghimi2016} apply GD-MC Boosting \citep{Saberian2011} to CNN and show that it is preferable to bagging for ensemble learning with CNNs. In \citep{Mosca2017} authors argue that random initialization of the network at each round of boosting is unnecessary; they advocate weight transfer from the previous boosting step. Liu et al. \citep{Liu2018} use to label data for online relearning of strong classifiers cascade with Haar-features
	\label{sec:headings}
	
	\subsection{Dynamically configurable neural networks}
	A lot of research studies ways to extend neural networks with effective dynamically configured computation graph. One of the objectives is to save computational resources by distinguishing hard and easy examples. Graves \citep{Graves2016} modifies recurrent neural network architecture to allow adaptive computational time (ACT). Figurnov et al. \citep{Figurnov2016} used ACT in residual blocks of convolutional neural networks and applied them for object detection. In \citep{McGill2017} the network decides if it continues to process an image with “stop” and “go” signals. The classification process is encapsulated in a single network architecture that shares the internal representation of the individual sub-modules. In contrast to our approach, the “classifier selection function” (defined in section \ref{section:la_class}) cannot be learned explicitly in all the cases.
	
	In \citep{Neshatpour2018} several separate networks of different sizes are launched successively; the classification stops on arbitrary step based on the estimated confidence. Each network takes a different sub-band generated from a Discrete Wavelet Transformation of an input image. First networks operate with coarser resolution, therefore, consume fewer computation resources than the next ones. A similar “coarse-to-fine analysis” effect can be achieved by fast saccadic moves in the mechanism of hard visual attention, which can be learned via reinforcement learning. The first work in that direction is Recurrent visual attention (RAM) \citep{Mnih2014}; at each timestep, an agent observes only a part of an image and controls the eyesight direction to concentrate on the most informative regions. \citep{Liu2018} has shown that RAM can be improved with dynamic computational time (DT-RAM) by providing the network ability to produce a stop signal; on average DT-RAM needs fewer steps to provide the same or better results on MNIST. In \citep{Bellver2016} and \citep{Wang2017} an agent learns to control not only position but also the size of the window that enables to focus on objects of different sizes. Additionally, in \citep{Wang2017} the agent observers VGG feature space, instead of raw pixels. Hard visual attention, however, doesn’t imply branching of the internal structure of computations, which is the goal of our research.
	
	Conceptually close to the least action principle are image-enhancement networks with dynamically configurable computations \citep{yu2018, yu2019}. Their key idea is that some parts of the image are uniform and easier to denoise and, therefore, should be processed differently from silent ones. Yu et al.\citep{yu2018, yu2019} adapt reinforcement learning and train different toolchains that the agent can use. In \citep{Huang2017} authors similarly teach an agent to skip layers of neural networks in the task of visual object tracking.
	
	Recently self-attention mechanism provided by transformers shows promising results when applied to computer vision problems \citep{Dosovitskiy2020, Carion2020}, though these works concentrate on performance benefits, do not adapt hard attention; the least-action principle is ignored.
	
	\subsection{Meta learning}
	The problem of learning a policy to select an algorithm from a list is known as algorithm selection (AS) task \citep{Rice76}. Recently introduced Dynamic Algorithm Configuration (DAC) \citep{DAC2020} in contrast to conventional AS suggests exploiting iterative nature of real tasks when an agent has to reveal important details of concrete example iteratively. Biedenkapp et al \citep{DAC2020} formulate the problem as contextual Markov Decision Process (contextual MDP), underlying the fact that context plays a crucial role in an exact configuration. They show that reinforcement learning is a robust candidate to obtain configuration policies: it outperforms standard parameter optimization approaches. 
	
	Sparse ensemble learning itself exploits an iterative nature, therefore, our approach can be thought of as a special case of DAC: context is a concrete image, reward takes into account computational savings, action and observation spaces have particular forms. These aspects lead to a different agent architecture and loss function than in \citep{DAC2020}. Our setup also relates to contextual bandits, but instead of just learning the action selection policy, an algorithm has to interpret the bandit (classifier) response.
	
	In this work, we create the set of classifiers that are useful for the agent, instead of learning a single classifier with sub-modules inside. Introducing non-differential operations might look like excessive complications because the supervised signal is richer and the training procedure is simpler. But with separate modules we can control that the optimal policy has to exploit dynamically configured computations, and “selection function” is better learned via reinforcement signal \citep{Mnih2014}.
	
	Sparse ensemble learning allows a seamless iterative increase of complexity without retraining from scratch, because agent itself can be treated as a “tool”; this may help to create systems that gradually become more complex.
	
	We see our contribution as two-fold:
	\begin{enumerate}
		\itemsep=3pt
		\item We state sparse ensemble learning problem based on the least-action principle as a special case of Dynamic Algorithm configuration.
		\item We propose a Least Action Classifier network architecture and appropriately designed loss function to solve the stated problem. We show by experiment, Least Action Classifier has the advantage over conventional ensemble learning (stacking) under computational expenses restrictions.
	\end{enumerate}
	
	\section{Least action classification}
	\label{section:la_class}
	In previous work \citep{Malashin2019} we showed that computationally efficient ensemble of classifiers under some assumptions has to implement two key functions:
	\begin{enumerate}
		\item Classifier selection function $ \Phi_1: S^{(t)} \mapsto \hat{a}^{(t)}$.
		
		\item State update function $\Phi_2 : \{S\textsuperscript{(t)},y^{(t)}\}  \mapsto S^{(t+1)}$.
	\end{enumerate} 
	In the case $ S $ is an internal (hidden) representation of the current task status at step $ t $, $ y^{(t)} $ is selected classifier response, $ \hat{a}^{(t)} $ is classifier “key”(index).
	Classifier selection function takes state as input and produces “key” of a classifier in the pool. The state update function purpose incorporates knowledge about classifier response in the state representation. 
	
\begin{figure}[!h]
		\centering
		\includegraphics[scale=.21]{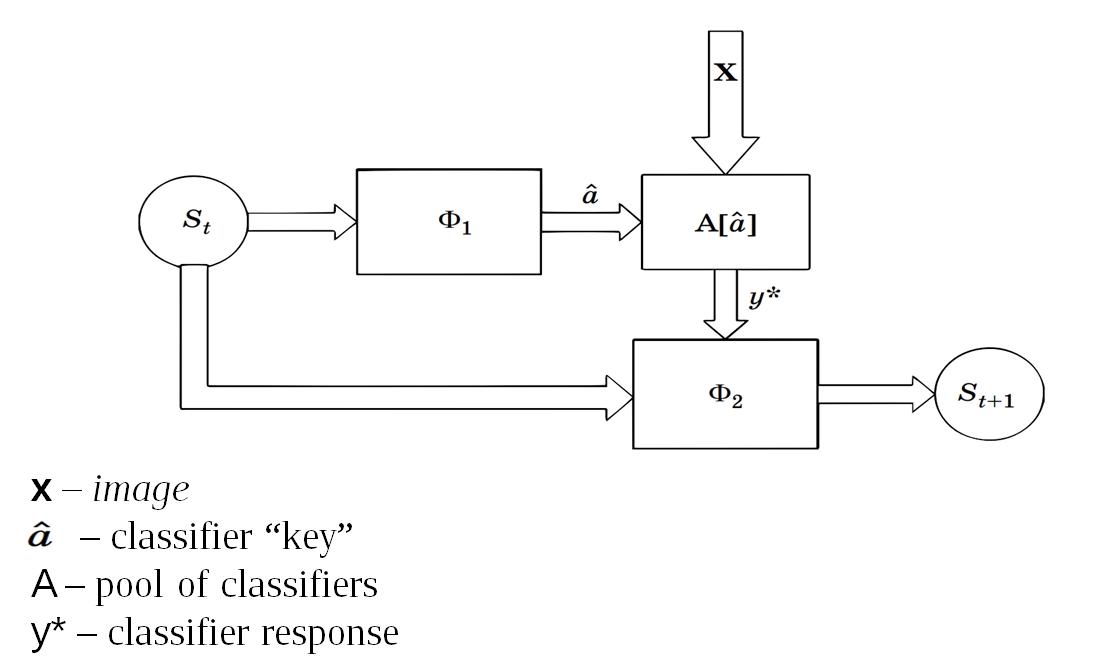}
		\caption{Markov decision process diagram \citep{Malashin2019}}
		\label{fig:MDP_LAC}
	\end{figure}

	The problem might be represented by Markov decision process shown in Figure \ref{fig:MDP_LAC} \citep{Malashin2019}.
	
	Due to mutual dependency of classifiers responses, finding optimal $\Phi_1$ and $\Phi_2$ is hard as the original classification problem, but approximations can be learned in the reinforcement learning setup with the following elements:
	\begin{enumerate}
		\item Environment is a) an image and b) the pool of classifiers.
		\item The action space consists of a) indexes of the classifiers in the pool and b) the prediction (label or probability distribution over classes).
		\item Observation is the responses of the classifiers.
		\item Episode is classifying a single image.
		\item The reward takes into account the accuracy of the decision and computational complexity of the selected classifiers.
	\end{enumerate}
	
	We can think about the approach as “sparse” stacked generalization \citep{Wolpert1992} when meta learner is an agent and base models are CNNs. We want the learner to assign zero weights for most of the classifiers' predictions, but exact “zeros” are discovered iteratively and individually for every image. Across the sample distribution, computationally heavy classifiers should be zeroised more often than lightweight classifiers. This might improve generalization because large models often tend to overfit.
	
	\subsection{Pool of classifiers}
	To learn agent policies, we need to create an initial pool of classifiers, through which an agent will be able to interact with an image. Intuitively, the desirable properties of the classifiers are decorrelated responses and computational exuberance of architectures. We consider two classifier types:
	\begin{enumerate}
		\item CNNs learned by iteratively increasing weights of the images that were incorrectly classified on the previous step (boosting).
		\item CNNs learned on different class subsets. These classifiers provide good variability of network responses.
	\end{enumerate}
	
	\paragraph{Boosting}
	The goal of conventional boosting is to ensemble a committee $ f $, which has the following form:
	\begin{equation}
	\label{eq:boosting_committee_form}
	f(\textbf{x}) = \sum_{m=1}^{M}w_m f_m(\textbf{x}), 
	\end{equation}
	where $f_m$ is the \textit{m}-th weak learner and $w_m$ is its weight.
	
	Boosting implicitly assume that different classifiers from the committee concentrate on different examples. Therefore, $\{ f_m \}$ can provide a good environment for an agent that has the goal to learn a policy that avoids using every classifier for every image. We implemented BoostCNN \citep{Moghimi2016} that carries the optimization by gradient descent in the functional space with the GD-MC approach.
	
	Moghimi et al show that GD-MC is preferable to bagging for CNNs. But according to our experiments, the advantage of BoostCNN in their experiments on CIFAR-10 can be explained by under-fitting individual networks during single bagging iteration. We optimized some parameters and concluded that bagging outperforms BoostCNN in this task. Only when there are very few boosting iterations (e.g., 2), BoostCNN sometimes provides a better committee. We give more details in appendix A.
	
	We have also experimented with Multi-Class Adaboost SAMME \citep{zhu2009} that re-weights training examples after each iteration of boosting. SAMME supports arbitrary loss function (not only with mean-squared error), including cross-entropy, commonly used for classification. But experiments showed that the weighted learning procedure converges badly for CNNs because of large variance across weights after each boosting iteration. One can solve the issue by forming the training set according to boosting weights (Adaboost.M2), but we did not explore this approach.
	
	In \citep{Mosca2017} authors suggest a successive increase of networks’ depth on each iteration of boosting. We tried to extend the approach by freezing weights obtained on the previous iteration of boosting. In this case features of classifier on the $ t-1 $-th iteration of boosting can be used without re-computation in the deeper classifier $ t $. However, we observed that without fine-tuning all the layers, the accuracy of the committee does not improve from iteration to iteration.
	
	We experimented with the idea of underfitting networks in the first boosting iteration, and the impact was inconsistent.
	
	Therefore, in our experiments, the simple bagging approach outperforms the conventional boosting of CNN classifiers. At the same time classifiers obtained in bagging lack specificity that we need to study agent ability to produce context-dependent sequence of actions.
	
	\paragraph{Classifiers trained with different class subsets}
	The approach of learning classifiers on different subsets of classes guarantees the specificity and (at least partial) decorrelation of responses. As a negative consequence reducing the space of recognized classes causes poorer gradients \citep{Malashin2016} and therefore harms training. For research purposes however different “task” forces classifiers to have less correlated responses. For large problems specificity of different modules can arise naturally.
	
	Let the dataset $ D $ consist of $ N $ images $ x_i $ with appropriate labels $ y_i $:
	\begin{equation}
	D = \{(x_i,y_i), i \in [1, N], x \in X, y \in Y\}.
	\end{equation} 
	Subsets of classes $ Y^k \subset Y $ split $D$ into overlapping datasets $ D_k $:
	\begin{equation}
	D_k = \{(x_k,y_k) \in D, y_k \in Y^k\}.
	\end{equation} 
	
	Separate classifiers learned on every $D_k$ form the pool of classifiers.
	
	\subsection{Least action classifier}
\begin{figure}[!h]
		\centering
		\includegraphics[scale=.25]{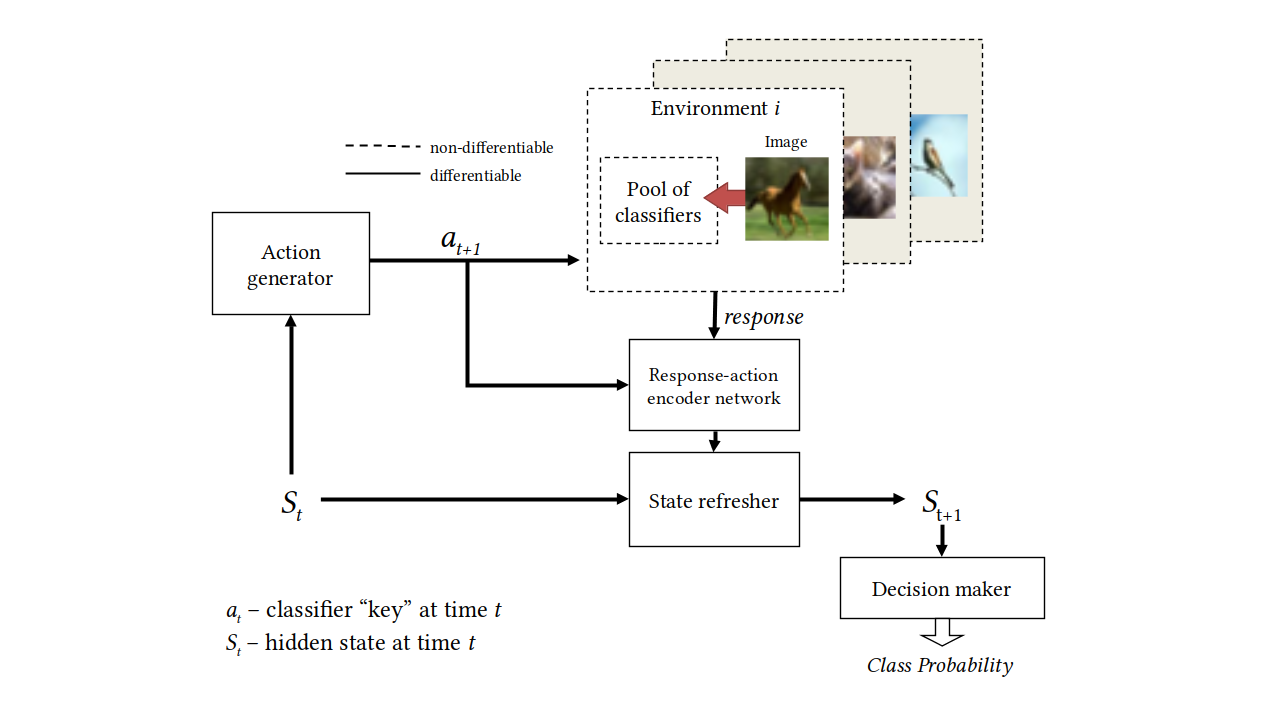}
		\caption{Least action classifier network architecture. Blocks of solid lines are differentiable. Sample images are from dataset CIFAR-10, which is collected by Alex Krizhevsky, Vinod Nair, and Geoffrey Hinton[MIT License](https://www.cs.toronto.edu/~kriz/cifar.html)}
		\label{fig:NET_LAC}
	\end{figure}

	Neural networks can be good candidates to approximate functions $\Phi_1$ and $ \Phi_2 $. We come up with the Least-Action Classifier (LAC) depicted in Figure \ref{fig:NET_LAC}.
	
	LAC consists of the following five main components:
	\begin{enumerate}
		\item Environment response generator, a non-differentiable element that takes an image and the index of the requested classifier, and returns response of the classifier.
		\item State refresher that implements $ \Phi_2 $ function; at step $ t $ it inputs hidden state vector and encoded classifier response; returns new hidden state vector.
		\item Action generator that implements $\Phi_1$ function; it inputs hidden state vector and returns the “key” of the classifier.
		\item Decision maker that inputs hidden state vector and outputs current solution.
		\item Response-action encoder that encodes action and classifier response in the format that is appropriate for state refresher.
	\end{enumerate}
	
	LAC architecture is flexible in the selection of its components. For example, some existing architectures of visual attention can be implemented as LAC classifiers by replacing an action generator with a location policy network. The key difference is that LAC uses explicitly learned CNN classifiers, which can be deeper than ones learned via reinforcement learning.
	
	State refresher can be represented by LSTM or other type of recurrent network, but in our experiments, it appeared that more robust dynamically configured computations are learned with state refresher with a short memory. We refer to LAC with a state refresher of this type as LAC-sm. Figure 4 shows its hidden state structure and diagram of the update algorithm.
	
	\begin{figure}[!h]
		\centering
		\includegraphics[scale=.30]{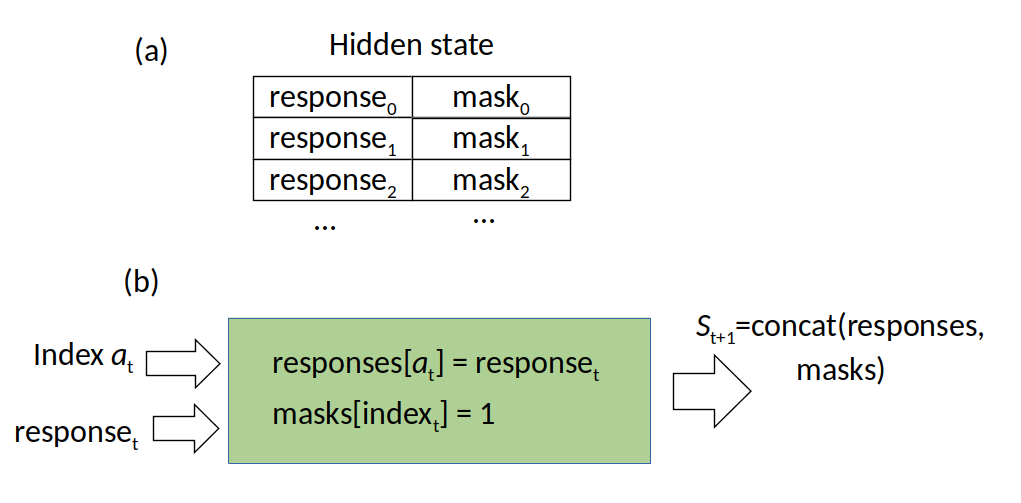}
		\caption{ State refresher with short memory, Hidden state structure (a) and the diagram of its update algorithm (b)}
		\label{fig:LAC_sm}
	\end{figure}
	
	Instead of encoding the state with hidden units, we store responses of every classifier in a table. Additionally, we extend the state representation with a table of masks that we set to ones when saving the corresponding classifier response into the first table. The masks provide a clear marker of what classifiers had been already called; they help to avoid duplicate actions. Two tables comprise a hidden state. The size of a state-vector is, therefore, $N \times C \times 2$, where $N$ is a number of responses to memorize and $ C $ is the size of the response vector. At the beginning of an “episode”, both tables are filled with zeros. 
	
	Response-action encoder for LAC-sm is the identity mapping of classifier response and classifier index. LAC-sm does not need to have recurrent connections at all as soon as the memory is hardwired in a non-differentiable manner.
	
	The action generator structure consists of two fully connected layers with RELU activation. It returns the probability of the classifiers to be called on the next step. The decision maker has three fully connected layers with RELU and returns probability distribution over image classes.
	
	\subsection{Loss function}
	
	Similar to recurrent visual attention model \citep{Mnih2014} LAC is learned by hybrid function:
	
	\begin{equation}
	{Loss} = \gamma L_{RL} + {Loss}_{S},
	\end{equation}
	where $ L_{RL} $ refer to reinforcement loss, and $ {Loss}_{S} $ refers to standard cross-entropy loss (with ground truth label), $ \gamma $ is hyperparameter (we use $ \gamma $=0.01 in our experiments). We apply intermediate supervision \citep{Li2017} by computing supervised loss on every step of an episode. Reinforcement loss is a sum of action loss $ L_{action} $ and entropy-bonus $ L_H $:
	
	\begin{equation}
	{Loss}_{RL}=L_{action}+\alpha L_H
	\end{equation}
	
	where $ \alpha $ is a hyper-parameter (we use $\alpha=0.5$) and $ L_{action} $ has the following form:
	
	\begin{equation}
	\label{eq:action_loss}
	L_{action} = \sum_{k}^{K} \sum_{t}^{T} A_{k,t} log(\pi(a_{k,t} | s_{k,t-1};\theta_a)], 
	\end{equation}
	
	where $K$ is the number of images in the batch, $T$ is the number of actions taken in each “episode”, $\pi$ is action policy, $ \theta_a $ is vector of action generator’s weights,	$ A_{k,t}= R_{k,t}-b(sk,t-1) $ is advantage, an extra reward $ R $ over prediction of the baseline network $ b $ agnostic to the action taken. In our experiments only shallow one-layer baseline networks provided learning policy with dynamically configured computations. We have found that alternatively, we can use a deeper two-layer network with dropout.
	
	Formula \ref{eq:action_loss} refers to A3C loss because batches of images are analogous to multiple environments.
	
	Entropy bonus has the following form:
	
	\begin{equation}
	\label{eq:ent_bonus}
	L_H = \sum_{i=1,t=2}logP(a_{i,t})P(a_{i,t}) + \beta \sum_{k=1,t=1}logP(a_{k,i,t})P(a_{k,i,t}),
	\end{equation}
	where $ \beta $ is hyperparameter, $ P(a_{i,t}) $ is probability of selecting classifier $ i $ on step $ t $ averaged across all $ K $ images in a batch. To force the agent to use different classifiers on different steps, in the first term of (\ref{eq:ent_bonus}) we use entropy of actions chosen in course of every episode, starting with second step, because first step is context-free. Second term softens predicted action distribution avoiding non-alternative decisions during training. In the experiments $ \beta =10^{-4}$.
	
	The reward for every episode has the form:
	
	\begin{equation}
	R = r - \lambda \sum_{i \in [1,c]} T(a_i),
	\end{equation}
	
	where $ r $ equals 1 if the image is classified correctly and 0, otherwise, $ T(a_i) $ is the time needed to execute a classifier associated with action $ a_i $, $ \lambda \ge 0 $ is hyperparameter, and $ c $ is a number of classifiers that the agent used before producing the final response. In the experiments we used fixed $ c $ that is less than a number of all classifiers in the pool, therefore, we assumed that $ \lambda=0 $. 
	
	\section{Experiments}
	In experiments, we used CIFAR-10, which has 50000 train and 10000 test $ 32 \times 32 $ color images of 10 object classes.
	
	\subsection{Pool of classifiers}
	We used two simple CNN architectures in our experiments. First has two constitutional layers with 6 and 16 filters, followed by three fully connected layers with 120, 84, and 10 neurons respectively. Each convolutional layer is followed by max-pooling. The second architecture has no fully-connected layers. It consists of three convolutional layers with max-pooling (after 1-st layer) and average pooling (after 2-nd and 3-rd layers). RELU activation is everywhere except the top of the networks.
	We made a random search on learning parameters and used them for every CNN network in our environment. The best results on average were obtained with SGD optimizer, geometric augmentation, batch size of 128 and step learning schedule with a start rate of 0.01, decreasing in the course of training. In Table~\ref{tab:pool1} there are the six classifiers we learned on randomly chosen subsets of 10 original CIFAR-10 classes; we chose net architecture for a classifier randomly as well.
	
	\begin{table}
		\caption{Pool 1 of CNN classifiers learned on a subset of image classes of CIFAR-10 dataset}
		\centering
		\begin{tabular}{llll}
			\toprule
			\# & Image classes & Arch. type	& Test acc (10 classes)		\\
			\midrule
			0 	& \{0,1,8,4\}   		& 1 	& 35.6 			\\
			1 	& \{1,2,3,5,6,7,9\} 	& 2 	& 57.09 		\\
			2 	& \{3,2,4\} 			& 2 	& 24.64 		\\
			3 	& \{7,2\} 				& 2 	& 18.26 		\\
			4 	& \{0,1,6,7,8,9\} 		& 1 	& 51.02 		\\
			5 	& \{0,2,3,5\} 			& 1 	& 29.49 		\\
			\bottomrule
		\end{tabular}
		\label{tab:pool1}
	\end{table}
	
	\subsection{Sparse ensemble learning}
	We train LAC for 200 epochs with Adam optimizer. The learning rate is decreased by the factor of 10 after epochs 170 and 190.
	
	In the first experiment we threshold the number of actions for LAC-sm. Table \ref{tab:lac_actions} shows the results.
	
	\begin{table}[h!]
		\caption{LAC performance on CIFAR-10 with the different number of actions in pool 1}
		\centering
		\begin{tabular}{llll}
			\toprule
			Method & Accuracy, \% 	\\
			\midrule
			averaging all responses & 62  	\\
			LAC-sm with 1 action 	& 67.8 \\
			LAC-sm with 2 action 	& 75.81 \\
			LAC-sm with 3 action 	& 77.81 \\
			LAC-sm with 4 action 	& 78.62 \\
			LAC-sm with 5 action 	& 79.1 \\
			LAC-sm with 6 action 	& 79.29 \\
			\bottomrule
		\end{tabular}
		\label{tab:lac_actions}
	\end{table}

	We conclude that the agent can incorporate information from multiple classifiers, however, it is not clear if the agent learns efficient context-dependent classifier selection function $\Phi_1$. 
	
	We compare LAC with a context-agnostic baseline to verify this. First, we have found the most appropriate algorithm for stacking responses of the classifiers from the pool. Among different machine learning algorithms, a neural network with 5 fully-connected layers produced the best result (79,5\% of accuracy), which is slightly better than the Least action classifier with six actions. In the experiments below we used a shallower multilayer perceptron (MLP) with 3 fully-connected layers as a baseline. It provided almost the same result, being almost twice as small in the number of free parameters. The results of all other methods are in the appendix. 
	
	For the next experiment, we form pool 2 by selecting classifiers (with indexes 0,2,3,5) that complement each other in the data they were trained on. Then we train the baseline on every combination of the classifiers in pool 2 and compare it with LAC in Table \ref{tab:lac_pool2}.

	\begin{table}[h!]
		\caption{Accuracy on CIFAR-10 with the usage of classifiers from pool 2}
		\centering
		\begin{tabular}{lll}
			\toprule
			Number of classifiers used & MLP (best combination) & LAC-sm	\\
			\midrule
			4 	& 72.4	& \textbf{72.9} 	\\
			3 	& 69.7	& \textbf{71.6} \\
			2 	& 66.3	& \textbf{68.1} \\
			1 	& 59.8	& \textbf{60.0} \\
			\bottomrule
		\end{tabular}
		\label{tab:lac_pool2}
	\end{table}
	
	As expected, exclusion of any classifier drops the accuracy providing resource vs accuracy conflict in the pool. Table \ref{tab:lac_pool2} shows that under computational restrictions agent learns to dynamically adapt to the image content and can negate the drop of accuracy by a large margin.
	
	On the test set LAC-sm with four allowed actions (LAC-sm-4) uses every classifier evenly, while LAC-sm-1 uses only the best one. These policies are naturally context-independent and were expected to provide the same results as the baseline.
	
	Surprisingly  LAC-sm-4  outperforms baseline by more than  0.5\%.  One explanation is that intermediate supervision and noisy training provides a dropout-like regularization effect by forcing decision maker to guess in an absence of some responses.
	
\begin{figure}[!h]
		\centering
		\includegraphics[scale=.25]{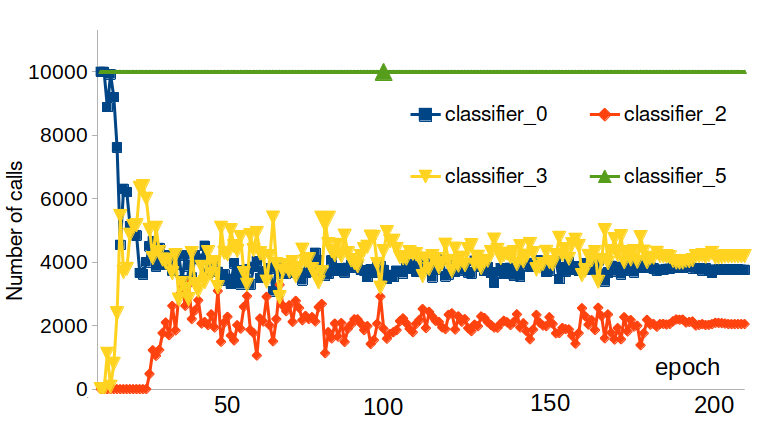}
		\caption{Classifier call frequency on test set during course of training of LAC-sm-2}
		\label{fig:class_freq}
	\end{figure}

	However, we have evidence that LAC-sm-2  and  LAC-sm-3 have learned context-dependent policy: they outperform baseline significantly. Figure~\ref{fig:class_freq}~ shows that LAC-sm-2 uses every classifier with a different frequency, which shows its ability to exploit context.
	
	Finding the best combination of classifiers is easy for LAC, but revealing good context-dependent policy often takes many epochs. Figure 5 shows that until twentieth epoch the agent ignored classifier 2.
	
	In our experiments, dynamic computations are the key factor to produce delta in accuracies of the Least Action Classifier and the baseline shown in Table \ref{tab:lac_pool2}. Figure 6 depicts the computational graphs of two versions of LAC-sm-3 trained with different parameters.
	
	\begin{figure}[!h]
		\centering
		\includegraphics[scale=.25]{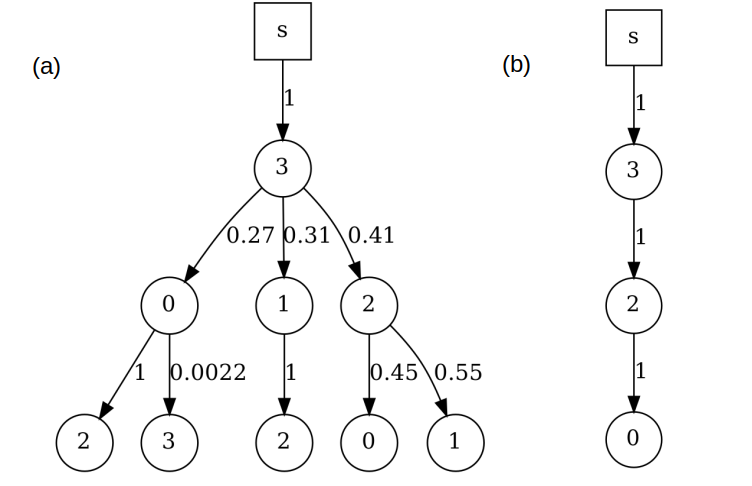}
		\caption{Diagram representing graph of computations. Edges are probabilities, nodes are classifiers, "s" node refer to the start. (a) LAC-sm-3 trained with entropy bonus and 128 units in decision maker's hidden layers (got 71.6\% accuracy on the test set), (b) LAC-sm-3 trained without entropy bonus and 128 units in decision maker's hidden layers (69.5\% accuracy)}
		\label{fig:dyn_graphs}
	\end{figure} 
	
	Without entropy bonus and excessively large decision maker,  Least  Action  Classifier learns computational graph shown in Figure \ref{fig:dyn_graphs}b, which incorporates only a single trajectory; it simply ignores classifier \#1. The resulted test accuracy is on par with the context agnostic baseline. With proper parameters, LAC exploits five different trajectories (Figure \ref{fig:dyn_graphs}a) and outperforms baseline by almost 2\%.
	
	\section{Conclusion}
	
	In this work, we formulate CNN sparse ensemble learning problem when an agent is taught to incorporate knowledge from several pre-trained classifiers taking into account their computational complexity. The goal of the agent is to learn context-dependent policy to unfold a computational graph in a way that refers to maximum expected accuracy under condition of limited number of actions. We introduce Least action classifier architecture with a short memory and an appropriate loss function. We show by experiment that Least action classifier learns a policy that outperforms the conventional approach of stacking CNN classifiers. Sparse ensemble learning allows a seamless iterative increase of complexity without retraining from scratch, because agent itself can be treated as a “tool”; this may help to create systems that gradually become more complex.
	
	\section{Acknowledgments}
	This research was funded by Russian Science Foundation (project 19-71-00146).
	
	\bibliographystyle{unsrtnat}
	\bibliography{references}

\begin{thebibliography}{27}
\providecommand{\natexlab}[1]{#1}
\providecommand{\url}[1]{\texttt{#1}}
\expandafter\ifx\csname urlstyle\endcsname\relax
  \providecommand{\doi}[1]{doi: #1}\else
  \providecommand{\doi}{doi: \begingroup \urlstyle{rm}\Url}\fi

\bibitem[Malashin(2019)]{Malashin2019}
R.~Malashin.
\newblock Principle of least action in dynamically configured image analysis
  systems.
\newblock \emph{J. Opt. Tech.}, 86, 2019.

\bibitem[Shelepin and Krasilnikov()]{Shelepin2003}
Y.E. Shelepin and N.N. Krasilnikov.
\newblock Principle of least action, physiology of vision and conditioned
  reflex theory.
\newblock \emph{Ross. Fiziol. Zh. im. I. M. Sechenova}, 89\penalty0 (6).

\bibitem[Shelepin et~al.(2006)Shelepin, Krasilnikov, Trufanov, Harauzov,
  Pronin, and A]{Shelepin2006}
Y.~Shelepin, N.~Krasilnikov, G.~Trufanov, A.~Harauzov, S.~Pronin, and Foking A.
\newblock The principle of least action and viusal perception.
\newblock In \emph{Twenty-ninth European Conference on Visual Perception},
  volume~35, August 2006.

\bibitem[Viola and Jones(2001)]{ViolaJones2001}
P.~Viola and M.~Jones.
\newblock Rapid object detection using a boosted cascade of simple features.
\newblock In \emph{CVPR 2001}, volume~1, pages I--511--I--518 vol.1, 2001.
\newblock \doi{10.1109/CVPR.2001.990517}.

\bibitem[Murphy()]{Murphy2013}
Kevin~P. Murphy.
\newblock \emph{Machine learning : a probabilistic perspective}.
\newblock MIT Press, Cambridge, Mass. [u.a.].
\newblock ISBN 9780262018029 0262018020.

\bibitem[Mohammad~Moghimi and Li(2016)]{Moghimi2016}
Mohammad Saberian Jian Yang Nuno~Vasconcelos Mohammad~Moghimi, Serge~Belongie
  and Li-Jia Li.
\newblock Boosted convolutional neural networks.
\newblock In Edwin R.~Hancock Richard C.~Wilson and William A.~P. Smith,
  editors, \emph{Proceedings of the British Machine Vision Conference (BMVC)},
  pages 24.1--24.13. BMVA Press, September 2016.
\newblock ISBN 1-901725-59-6.
\newblock \doi{10.5244/C.30.24}.
\newblock URL \url{https://dx.doi.org/10.5244/C.30.24}.

\bibitem[Saberian and Vasconcelos(2011)]{Saberian2011}
Mohammad~J. Saberian and Nuno Vasconcelos.
\newblock Multiclass boosting: Theory and algorithms.
\newblock In J.~Shawe-Taylor, R.~S. Zemel, P.~L. Bartlett, F.~Pereira, and
  K.~Q. Weinberger, editors, \emph{Advances in Neural Information Processing
  Systems 24}, pages 2124--2132. Curran Associates, Inc., 2011.

\bibitem[Mosca and Magoulas(2017)]{Mosca2017}
Alan Mosca and George Magoulas.
\newblock Deep incremental boosting.
\newblock 08 2017.

\bibitem[Liu et~al.(2018)Liu, Li, Chang, and Dong]{Liu2018}
Chunsheng Liu, Shuang Li, Faliang Chang, and Wenhui Dong.
\newblock Supplemental boosting and cascaded convnet based transfer learning
  structure for fast traffic sign detection in unknown application scenes.
\newblock \emph{Sensors}, 18:\penalty0 2386, 07 2018.
\newblock \doi{10.3390/s18072386}.

\bibitem[Graves(2016)]{Graves2016}
Alex Graves.
\newblock Adaptive computation time for recurrent neural networks.
\newblock \emph{CoRR}, abs/1603.08983, 2016.
\newblock URL \url{http://arxiv.org/abs/1603.08983}.

\bibitem[Figurnov et~al.(2016)Figurnov, Collins, Zhu, Zhang, Huang, Vetrov, and
  Salakhutdinov]{Figurnov2016}
Michael Figurnov, Maxwell~D. Collins, Yukun Zhu, Li~Zhang, Jonathan Huang,
  Dmitry~P. Vetrov, and Ruslan Salakhutdinov.
\newblock Spatially adaptive computation time for residual networks.
\newblock \emph{CoRR}, abs/1612.02297, 2016.
\newblock URL \url{http://arxiv.org/abs/1612.02297}.

\bibitem[McGill and Perona(2017)]{McGill2017}
Mason McGill and Pietro Perona.
\newblock Deciding how to decide: Dynamic routing in artificial neural
  networks.
\newblock In Doina Precup and Yee~Whye Teh, editors, \emph{Proceedings of the
  34th International Conference on Machine Learning}, volume~70 of
  \emph{Proceedings of Machine Learning Research}, pages 2363--2372,
  International Convention Centre, Sydney, Australia, 06--11 Aug 2017. PMLR.
\newblock URL \url{http://proceedings.mlr.press/v70/mcgill17a.html}.

\bibitem[{Neshatpour} et~al.(2018){Neshatpour}, {Behnia}, {Homayoun}, and
  {Sasan}]{Neshatpour2018}
K.~{Neshatpour}, F.~{Behnia}, H.~{Homayoun}, and A.~{Sasan}.
\newblock Icnn: An iterative implementation of convolutional neural networks to
  enable energy and computational complexity aware dynamic approximation.
\newblock In \emph{2018 Design, Automation Test in Europe Conference Exhibition
  (DATE)}, pages 551--556, 2018.

\bibitem[Mnih et~al.(2014)Mnih, Heess, Graves, and kavukcuoglu]{Mnih2014}
Volodymyr Mnih, Nicolas Heess, Alex Graves, and koray kavukcuoglu.
\newblock Recurrent models of visual attention.
\newblock In Z.~Ghahramani, M.~Welling, C.~Cortes, N.~D. Lawrence, and K.~Q.
  Weinberger, editors, \emph{Advances in Neural Information Processing Systems
  27}, pages 2204--2212. Curran Associates, Inc., 2014.

\bibitem[Bellver et~al.(2016)Bellver, Giro-i Nieto, Marques, and
  Torres]{Bellver2016}
Miriam Bellver, Xavier Giro-i Nieto, Ferran Marques, and Jordi Torres.
\newblock Hierarchical object detection with deep reinforcement learning.
\newblock In \emph{Deep Reinforcement Learning Workshop, NIPS}, December 2016.

\bibitem[Wang et~al.(2017)Wang, Chen, Li, Xu, and Lin]{Wang2017}
Zhouxia Wang, Tianshui Chen, Guanbin Li, Ruijia Xu, and Liang Lin.
\newblock Multi-label image recognition by recurrently discovering attentional
  regions.
\newblock In \emph{IEEE International Conference on Computer Vision}, October
  2017.

\bibitem[Yu et~al.(2018)Yu, Dong, Lin, and Loy]{yu2018}
Ke~Yu, Chao Dong, Liang Lin, and Chen~Change Loy.
\newblock Crafting a toolchain for image restoration by deep reinforcement
  learning.
\newblock 04 2018.

\bibitem[Yu et~al.(2019)Yu, Wang, Dong, Tang, and Loy]{yu2019}
Ke~Yu, Xintao Wang, Chao Dong, Xiaoou Tang, and Chen~Change Loy.
\newblock Path-restore: Learning network path selection for image restoration.
\newblock \emph{arXiv preprint arXiv:1904.10343}, 2019.

\bibitem[Huang et~al.(2017)Huang, Lucey, and Ramanan]{Huang2017}
Chen Huang, Simon Lucey, and Deva Ramanan.
\newblock Learning policies for adaptive tracking with deep feature cascades.
\newblock pages 105--114, 10 2017.
\newblock \doi{10.1109/ICCV.2017.21}.

\bibitem[Alexey~Dosovitskiy()]{Dosovitskiy2020}
Alexander Kolesnikov Dirk Weissenborn Xiaohua Zhai Thomas Unterthiner Mostafa
  Dehghani Matthias Minderer Georg Heigold Sylvain Gelly Jakob Uszkoreit
  Neil~Houlsby Alexey~Dosovitskiy, Lucas~Beyer.
\newblock An image is worth 16x16 words: Transformers for image recognition at
  scale.
\newblock URL \url{https://arxiv.org/abs/2010.11929}.

\bibitem[Carion et~al.(2020)Carion, Massa, Synnaeve, Usunier, Kirillov, and
  Zagoruyko]{Carion2020}
Nicolas Carion, Francisco Massa, Gabriel Synnaeve, Nicolas Usunier, Alexander
  Kirillov, and Sergey Zagoruyko.
\newblock End-to-end object detection with transformers, 2020.

\bibitem[Rice(1976)]{Rice76}
John~R. Rice.
\newblock The algorithm selection problem.
\newblock \emph{Advances in Computers}, 15:\penalty0 65--118, 1976.
\newblock URL \url{http://dblp.uni-trier.de/db/journals/ac/ac15.html#Rice76}.

\bibitem[Biedenkapp1 et~al.(2020)Biedenkapp1, Eimer, Hutter1, and
  Lindauer]{DAC2020}
Andre Biedenkapp1, H.~Furkan~Bozkurt1andTheresa Eimer, Frank Hutter1, and
  Marius Lindauer.
\newblock Dynamic algorithm configuration:foundation of a new meta-algorithmic
  framework.
\newblock volume 163, 2020.

\bibitem[Wolpert(1992)]{Wolpert1992}
D~Wolpert.
\newblock Stacked generalization.
\newblock \emph{Neural Networks}, 1992.

\bibitem[Zhu et~al.(2009)Zhu, Rosset, Zou, and Hastie]{zhu2009}
Ji~Zhu, Saharon Rosset, Hui Zou, and Trevor Hastie.
\newblock Multi-class adaboost.
\newblock 2009.

\bibitem[Malashin(2016)]{Malashin2016}
R~Malashin.
\newblock Extraction of object hierarchy data from trained deep-learning neural
  networks via analysis of the confusion matrix.
\newblock \emph{J. Opt. Tech.}, 83, 2016.

\bibitem[Li et~al.(2017)Li, Yang, Liu, Wen, and Xu]{Li2017}
Zhichao Li, Yi~Yang, Xiao Liu, Shilei Wen, and Wei Xu.
\newblock Dynamic computational time for visual attention.
\newblock \emph{CoRR}, abs/1703.10332, 2017.
\newblock URL \url{http://arxiv.org/abs/1703.10332}.

\end{thebibliography}
	
	\cleardoublepage
	
	\begin{appendices}
		\section{BoostCNN}
		The goal of boosting is to solve the following optimization problem:
		\begin{equation}
		\label{eq:boosting_loss}
		f^*=\min_f R(f) = \min_f \sum_{i=1}^{N}L(y_i, f(\textbf{x}_i)),
		\end{equation}
		where $L(y,\hat{y})$ is some loss function, ${(x_i,y_i)}, i \in N$ are training samples and committee $f$ has the form (\ref{eq:boosting_committee_form}).

		Since the task of fitting the composite function is complex, boosting tackles the problem sequentially:
		\begin{equation}
		f_m(\textbf{x}) = f_{m-1}(\textbf{x}) + v \beta_m \phi(\textbf{x}; \theta_m),
		\end{equation}
		where $\theta_m$ are model parameters, $\beta$ is the weight minimizing (\ref{eq:boosting_loss}) and $0<v<1$ is shrinkage parameter. \\
		
		We implemented BoostCNN \citep{Moghimi2016} that carries the optimization by gradient descent in the functional space with the GD-MC approach.
		
		In the case loss function has the form:
		\begin{equation}
		\label{eq:gdmc_boost_loss}
		L(z_i, f(x_i))=\sum_{j=1,j\ne z} \exp\frac{1}{2}[<{y_z}_i, f(x_i)> - <y_j, f(x_i)>],
		\end{equation}
		where $z_i \in 1...M$ is class label and \textbf{y} is label code.
		
		According to gradient boosting methods CNN learns to replicate gradients of the objective function in functional space with MSE loss function; $\beta$ coefficient is found by the linear search minimizing (\ref{eq:gdmc_boost_loss}) along $f_m$ direction. Following \citep{Moghimi2016} we replaced linear search with the binary search of $\delta R$.
		
\begin{figure}[h!]
			\begin{subfigure}{.5\textwidth}
				\centering
				\includegraphics[width=.8\linewidth]{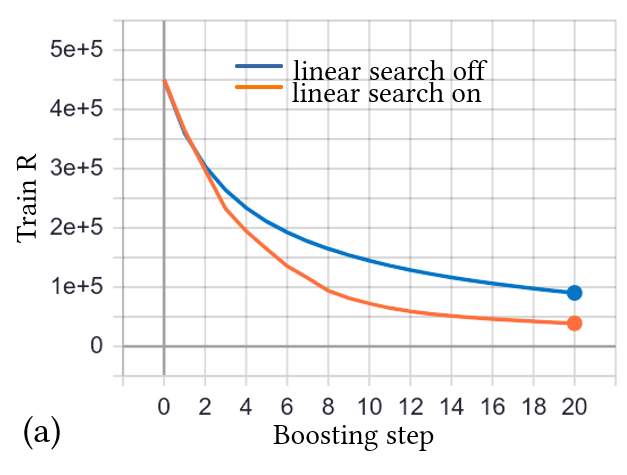} 
				\label{fig:sub-first}
			\end{subfigure}
			\begin{subfigure}{.5\textwidth}
				\centering
				\includegraphics[width=.8\linewidth]{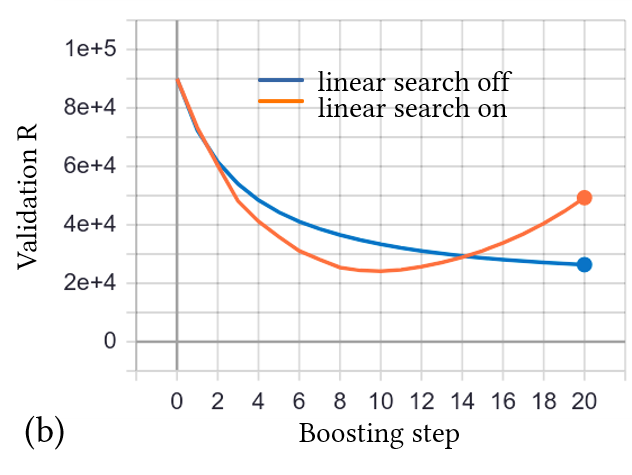} 
				\label{fig:sub-second}
			\end{subfigure}

			\begin{subfigure}{.5\textwidth}
				\centering
				\includegraphics[width=.8\linewidth]{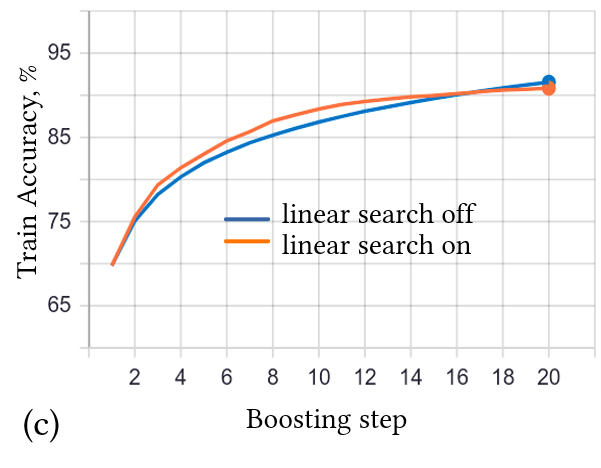} 
				\label{fig:sub-third}
			\end{subfigure}
			\begin{subfigure}{.5\textwidth}
				\centering
				\includegraphics[width=.8\linewidth]{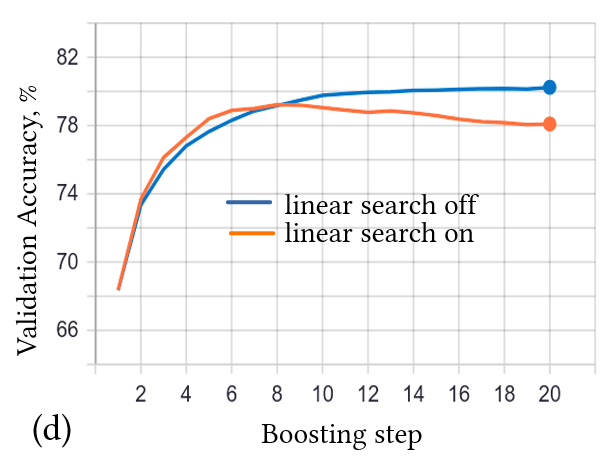} 
				\label{fig:sub-fourth}
			\end{subfigure}
			\caption{Learning curve for BoostCNN with and without linear search. (a), (b) - the objective function on the train and validation sets, bottom (c), (d) - the train and validation accuracy}
			\label{fig:gdmc_linsearch}
		\end{figure}

		We have found that the linear search impact on boosting process is ambiguous. We illustrate this on cifar-10 dataset. 
		For the first experiments, we reimplemented results from \citep{Moghimi2016} with cifar-quick network consisting of three convolutional layers with pooling and RELU activations and followed by two fully connected layers. Figure \ref{fig:gdmc_linsearch} shows training dynamics. 
		
		When using linear search shrinkage has appeared to have a significant impact, large $v$ can lead to unstable training of networks and sometimes the process can diverge due to excessive loss on heavily weighted examples. 
		As can be seen, linear search increases learning speed in first five boost steps but leads to overfitting after that.
		
\begin{figure}[h!]
			\begin{subfigure}{.33\textwidth}
				\centering
				\includegraphics[width=.8\linewidth]{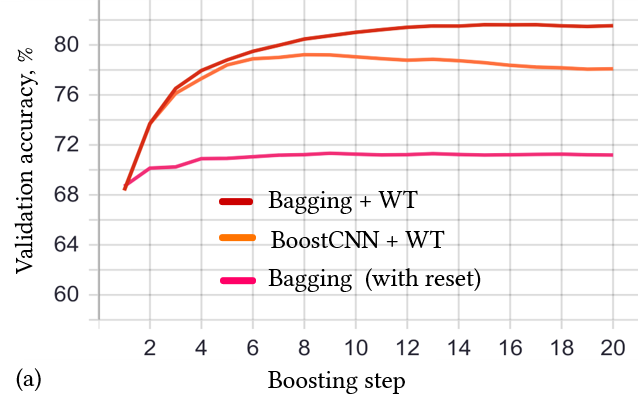} 
				\label{fig:gdmc_vs_bagging:a}
			\end{subfigure}
			\begin{subfigure}{.33\textwidth}
				\centering
				\includegraphics[width=.8\linewidth]{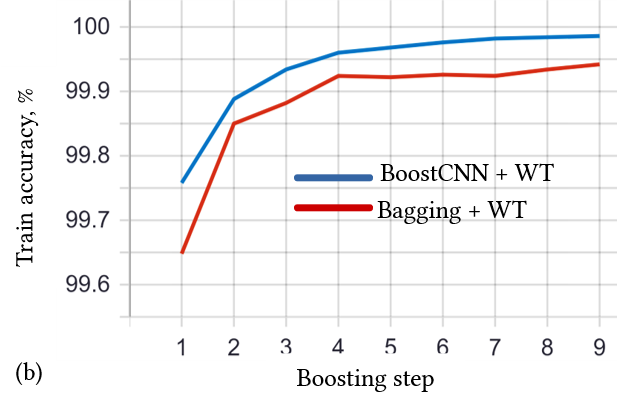} 
				\label{fig:gdmc_vs_bagging:b}
			\end{subfigure}
			\begin{subfigure}{.33\textwidth}
				\centering
				\includegraphics[width=.8\linewidth]{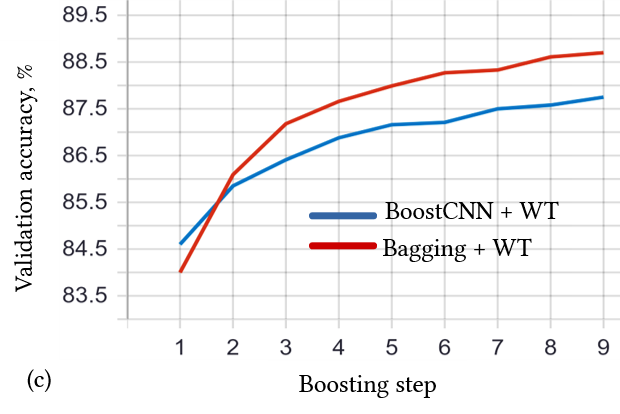} 
				\label{fig:gdmc_vs_bagging:c}
			\end{subfigure}
			\caption{Different ensemble learning methods on CIFAR dataset with simple CNN as base-learner. (a) committee validation accuracy of BoostCNN and bagging (with reset) reproduced with parameters from \citep{Moghimi2016} and comparison with bagging with weight transfer. (b) and (c) are learning curves for BoostCNN and bagging with weight transfer, resnet-18 architecture is a weak classifier, no augmentation is used during training}
			\label{fig:gdmc_vs_bagging}
		\end{figure}

		According to \citep{Moghimi2016} GD-MC is preferable to bagging for ensemble learning with CNNs and we were able to reproduce their results with the same network architecture used as base learner. However, according to our experiments, it can be argued that the advantage of BoostCNN over bagging in their experiments was achieved solely by under-fitting of individual networks during single bagging iteration. We optimized some parameters of Bagging: used a large bag for sampling with replacement (the same as the number of training examples), increased number of epochs per boosting step, added weight transferring (as \citep{Moghimi2016} did for GD-MC), and compared results with BoostCNN and unoptimized Bagging. Results are depicted on \figurename{\ref{fig:gdmc_vs_bagging}}a.
		
		One can see that bagging outperforms BoostCNN by more than 1\% in twenty iterations and (and is almost 3\% better than the result obtained in \citep{Moghimi2016}). Shrinkage tuning can slightly improve BoostCNN but still in our experiments it overfits to the tenth iteration while bagging improves further.  The conclusion  holds  for different  network architectures. Figure \ref{fig:gdmc_vs_bagging}b and Figure \ref{fig:gdmc_vs_bagging}c depict learning curves, when Resnet-18 is used as weak classifier. Bagging shows a much lower tendency to overfitting.
		
		One interesting finding is that (a) MSE loss on code words and (b) classical cross-entropy loss with one-hot encoding provide very similar training dynamics for individual networks. For example, resnet-18 trained for 100 epochs provided 7.5\% error rate with image augmentation and around 14\% without augmentation, no matter what loss function we used. Bagging with MSE and code-words provide slightly better results than bagging with cross-entropy and one-hot encoding.
		
		\cleardoublepage
		\section{Baseline}
		Table \ref{tab:baselines} shows the different methods' results in stacking responses of all classifiers from \tablename{} \ref{tab:pool1}. 
		
		\begin{table}[h]
			\caption{Stacking pool 1 classifiers with different methods }
			\centering
			\begin{tabular}{llll}
				\toprule
				Method & Test accuracy, \%	\\
				\midrule
				Adaboost + Decisions trees 	& 0.619			 \\
				SVM + rbf kernel			& 0.782			 \\
				SVM + linear kernel			& 0.79			 \\
				RandomForests				& 0.772			 \\
				ExtraTrees					& 0.756			 \\
				Decision Tree				& 0.677			 \\
				5-KNN						& 0.734			 \\
				15-KNN						& 0.755			 \\
				3-layer MLP					& 0.794			 \\
				\textbf{5-layer MLP	}				& \textbf{0.795} \\
				\hline
			\end{tabular}
			\label{tab:baselines}
		\end{table}
		
		Network with five fully connected layers and RELU activations got the best result. For generalization reasons in experiments, we preferred shallower three-layer MLP, which had given almost the same accuracy and at the same time has almost twice fewer parameters.
		
	\end{appendices}

\end{document}